\documentclass[10pt,twocolumn,letterpaper]{article}

\usepackage{arxiv}
\newtoggle{usetimes}
\newtoggle{itabs}
\toggletrue{usetimes}  

\iftoggle{usetimes}{\usepackage{times}}{}
\usepackage{amsmath,amssymb,amsfonts,dsfont,pifont,bm,bbm,mathrsfs,mathtools}
\usepackage{algorithm,algpseudocode,listings}
\usepackage{booktabs,multirow,adjustbox,diagbox,threeparttable}
\usepackage[pagebackref=false,breaklinks=true,colorlinks=true,citecolor=blue,bookmarks=false]{hyperref}
\usepackage{cleveref}  
\crefname{section}{Sec.}{Secs.}
\Crefname{section}{Section}{Sections}
\crefname{table}{Tab.}{Tabs.}
\Crefname{table}{Table}{Tables}
\crefname{figure}{Fig.}{Figs.}
\Crefname{figure}{Figure}{Figures}
\crefname{equation}{Eq.}{Eqs.}
\Crefname{equation}{Equation}{Equations}
\usepackage{wrapfig}

\usepackage{listings}
\definecolor{codegreen}{rgb}{0,0.6,0}
\definecolor{codegray}{rgb}{0.5,0.5,0.5}
\definecolor{codepurple}{rgb}{0.58,0,0.82}
\definecolor{backcolour}{rgb}{0.95,0.95,0.92}
\lstdefinestyle{mystyle}{
    backgroundcolor=\color{backcolour},
    commentstyle=\color{codegreen},
    keywordstyle=\color{magenta},
    numberstyle=\tiny\color{codegray},
    stringstyle=\color{codepurple},
    basicstyle=\ttfamily\scriptsize,
    breakatwhitespace=false,
    breaklines=true,
    captionpos=b,
    keepspaces=true,
    numbers=left,
    numbersep=5pt,
    showspaces=false,
    showstringspaces=false,
    showtabs=false,
    tabsize=2
}
\makeatletter
\newcommand{\defineRGBcolor}[2]{%
  \begingroup
  \def\color@values{\@gobble}%
  \@for\next:=#2\do{%
    \count0=\next\relax
    \multiply\count0 100
    \divide\count0 255
    \edef\color@values{\color@values,0.\number\count0}%
  }%
  \edef\x{\endgroup\noexpand\definecolor{#1}{rgb}{\color@values}}\x
}

\newcommand{\classaware}{\textcolor{black}{category-oriented channel awareness}\xspace}

\newcommand{\z}{{\rm\bf z}}
\newcommand{\x}{{\rm\bf x}}
\newcommand{\y}{{\rm\bf y}}
\newcommand{\e}{{\rm\bf e}}
\newcommand{\I}{{\rm\bf I}}

\renewcommand{\t}{t_e^c}

\begin{document}

\title{Interpreting Class Conditional GANs with Channel Awareness}

\author{Yingqing He$^\dagger$ \quad
    Zhiyi Zhang $^\ddagger$ \quad
    Jiapeng Zhu$^\dagger$ \quad
    Yujun Shen$^\ddagger$ \quad
    Qifeng Chen$^\dagger$ \\[5pt]
    $^\dagger$HKUST \quad
    $^\ddagger$ByteDance Inc.
}


\twocolumn[{
	\renewcommand\twocolumn[1][]{#1}
	\maketitle
	
	\defineRGBcolor{myred}{236, 122, 117}
    \defineRGBcolor{mygreen}{125, 148, 108}
    \defineRGBcolor{myblue}{138, 170, 243}
    \defineRGBcolor{myyellow}{216, 216, 108}
	\begin{center}
		\includegraphics[width=1.0\linewidth]{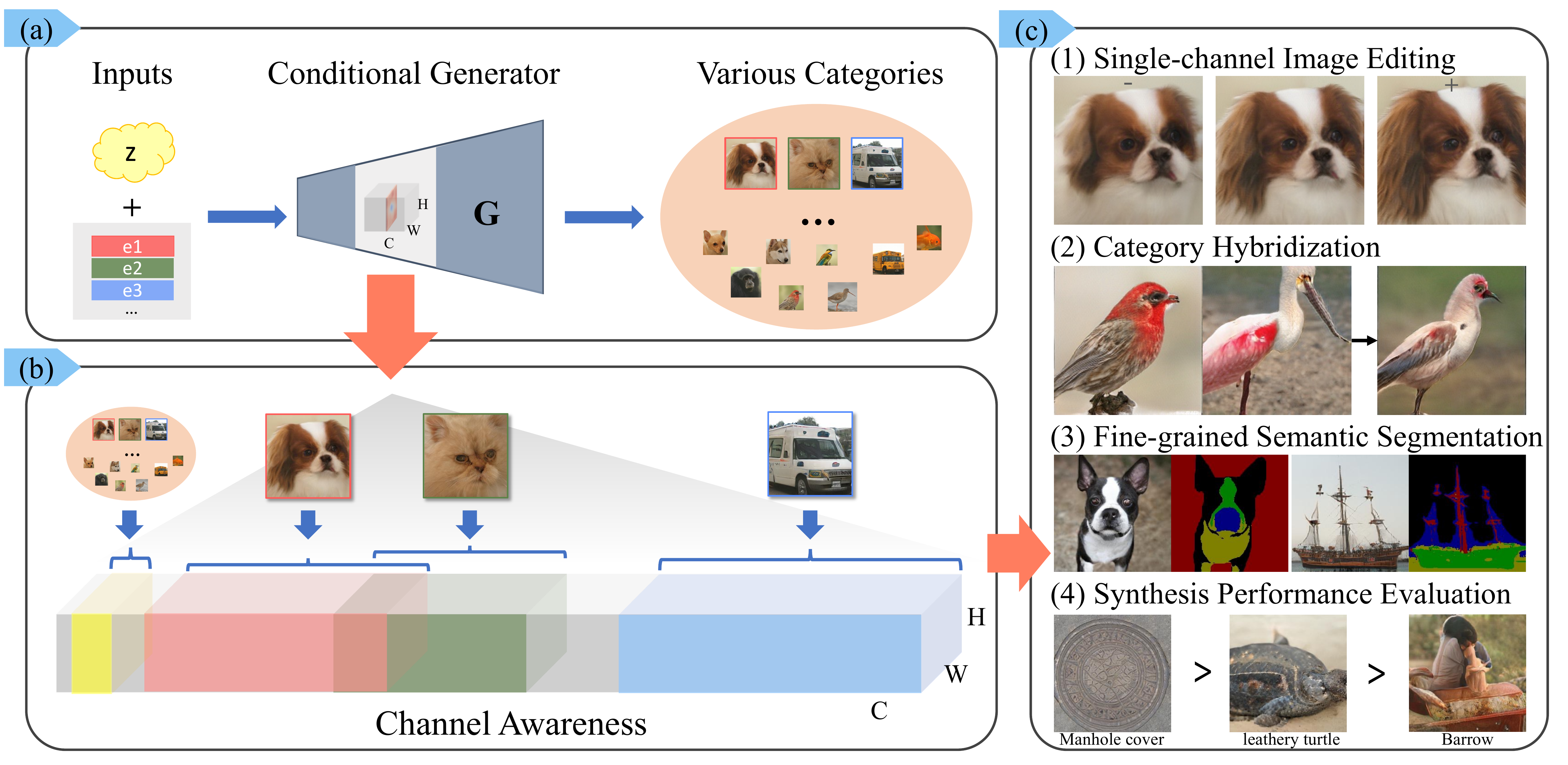}
		\vspace{-18pt}
		\captionsetup{type=figure}
		\caption{
            \textbf{Novel applications enabled by interpreting class conditional GANs.}
            Given a conditional generator in (a), we propose \textit{channel awareness} to quantify the contribution of each feature channel to the output image, as shown in (b), which reveals how the categorical information is handled by different channels.
            \textbf{\textcolor{myred}{Red}}, \textbf{\textcolor{mygreen}{green}}, and \textbf{\textcolor{myblue}{blue}} channels are primarily responsible for the synthesis of a particular category, while \textbf{\textcolor{myyellow}{yellow}} ones are shared by all classes.
            (c) Such an interpretation facilitates a range of applications, including single-channel image editing, category hybridization, fine-grained semantic segmentation, and category-wise synthesis performance evaluation.
        }
		\label{fig:teaser}
		\vspace{12pt}
	\end{center}
}]

\begin{abstract}

\iftoggle{itabs}{\it}{}

Understanding the mechanism of generative adversarial networks (GANs) helps us better use GANs for downstream applications.
Existing efforts mainly target interpreting unconditional models, leaving it less explored how a conditional GAN learns to render images regarding various categories.
This work fills in this gap by investigating how a class conditional generator unifies the synthesis of multiple classes. 
For this purpose, we dive into the widely used class-conditional batch normalization (CCBN), and observe that each feature channel is activated at varying degrees given different categorical embeddings.
To describe such a phenomenon, we propose \textit{channel awareness}, which quantitatively characterizes how a single channel contributes to the final synthesis.
Extensive evaluations and analyses on the BigGAN model pre-trained on ImageNet reveal that only a subset of channels is primarily responsible for the generation of a particular category, similar categories (\textit{e.g.}, cat and dog) usually get related to some same channels, and some channels turn out to share information across all classes.
For good measure, our algorithm enables several novel applications with conditional GANs.
Concretely, we achieve (1) versatile image editing via simply altering a single channel and manage to (2) harmoniously hybridize two different classes.
We further verify that the proposed channel awareness shows promising potential in (3) segmenting the synthesized image and (4) evaluating the category-wise synthesis performance.%
\footnote{Project page can be found \href{https://yingqinghe.github.io/interclassgan/}{here}.}

\end{abstract}
\vspace{-12pt}

\section{Introduction}\label{sec:intro}

The past few years have witnessed the rapid advancement of generative adversarial networks (GANs) in image synthesis~\cite{gan, pggan, stylegan, stylegan2, stylegan2ada, stylegan3, biggan}.
Despite the wide range of applications powered by GANs, like image-to-image translation~\cite{isola2017pix2pix, zhu2017unpaired, park2020contrastive}, super-resolution~\cite{ledig2017photo, wang2018esrgan, chan2021glean, menon2020pulse}, and image editing~\cite{park2020swapping, lample2017fader,ling2021editgan, bau2020semantic, patashnik2021styleclip}, it typically requires learning a separate model for a new task, which can be time and resources consuming.
Some recent studies have confirmed that a well-trained GAN model naturally supports various downstream applications~\cite{gu2020image, pan2020dgp, xu2021generative,chan2021glean,zhang2021datasetgan,2021linearsemantic, he2021unsupervised}, benefiting from the rich knowledge learned in the training process~\cite{bau2019gandissect, shen2020interfacegan, bau2020rewriting}.
Therefore, to make sufficient use of a GAN, it becomes crucial to explore and further exploit its internal knowledge.

Many attempts have been made to understand the generation mechanism of GANs~\cite{gansteerability, shen2020interfacegan, bau2019gandissect, 2021linearsemantic, wu2020stylespace, yang2019semantic, shen2020interpreting}.
It is revealed that, to produce a fair synthesis, the generator is required to render multi-level semantics, such as the overall attributes (\textit{e.g.}, the gender of a face image)~\cite{gansteerability, shen2020interfacegan}, the objects inside (\textit{e.g.}, the bed in a bedroom image)~\cite{bau2019gandissect, yang2019semantic}, the part-whole organization (\textit{e.g.}, the segmentation of the synthesis)~\cite{zhang2021datasetgan, 2021linearsemantic}, \textit{etc.}
However, existing efforts mainly focus on interpreting unconditional GANs, leaving conditional generation as a black box.

Compared with unconditional models, a class conditional model is more informative and efficient in that it unifies the synthesis of multiple categories, like animals, vehicles, and scenes~\cite{biggan}.
Figuring out how it manages the class information owns much great potential yet rarely explored.
To fill in this gap, we take a close look at the popular class-conditional batch normalization (CCBN)~\cite{biggan, sagan, sngan, contragan, projgan}, which is one of the core modules distinguishing conditional generators from unconditional ones.
Concretely, CCBN learns category-specific parameters to scale and shift the input features, such that the output features developed with different class embeddings can be easily told apart from each other, eventually resulting in the synthesis of various categories.
We notice from such a process that, under the perspective of the ReLU activation~\cite{relu} following CCBN, different feature channels present varying behaviors given different embeddings.

To quantify the aforementioned channel effect, we propose \textit{channel awareness} that characterizes how a single channel contributes to the final synthesis.
Through in-depth analyses on the BigGAN~\cite{biggan} model pre-trained on ImageNet~\cite{imagenet}, we have the following \textit{key findings}, which are also illustrated in \cref{fig:teaser}b.
First, only a portion of channels are active in rendering images for a particular class while the remaining channels barely affect the generation.
Second, more similar categories tend to share more relevant channels.
For instance, channels regarding dog synthesis intersect with those of cats but disjoint from those of buses.
Third, some channels highly response to the latent code instead of the class embedding and hence appear to deliver knowledge to all classes.

Beyond model interpretation, our proposed channel awareness facilitates a range of \textit{novel applications with class conditional GANs}, as shown in \cref{fig:teaser}c.
First, after identifying the relevant channels through awareness ranking, we realize versatile image editing by simply altering a \textit{single} feature channel (\cref{subsec:single_channel_manipulation}).
Second, through mixing the channels that are related to two classes respectively, we achieve harmonious category hybridization (\cref{subsec:channel_mixing}).
Third, we verify that intermediate feature maps from the generator, after weighted by our channel awareness, can be convincingly used for fine-grained semantic segmentation (\cref{subsec:segmentation}).
Fourth, we empirically demonstrate the potential of our channel awareness in evaluating the category-wise synthesis performance (\cref{subsec:total_channel_awareness}).

\begin{figure*}[t!]
    \centering
    \includegraphics[width=0.95\linewidth]{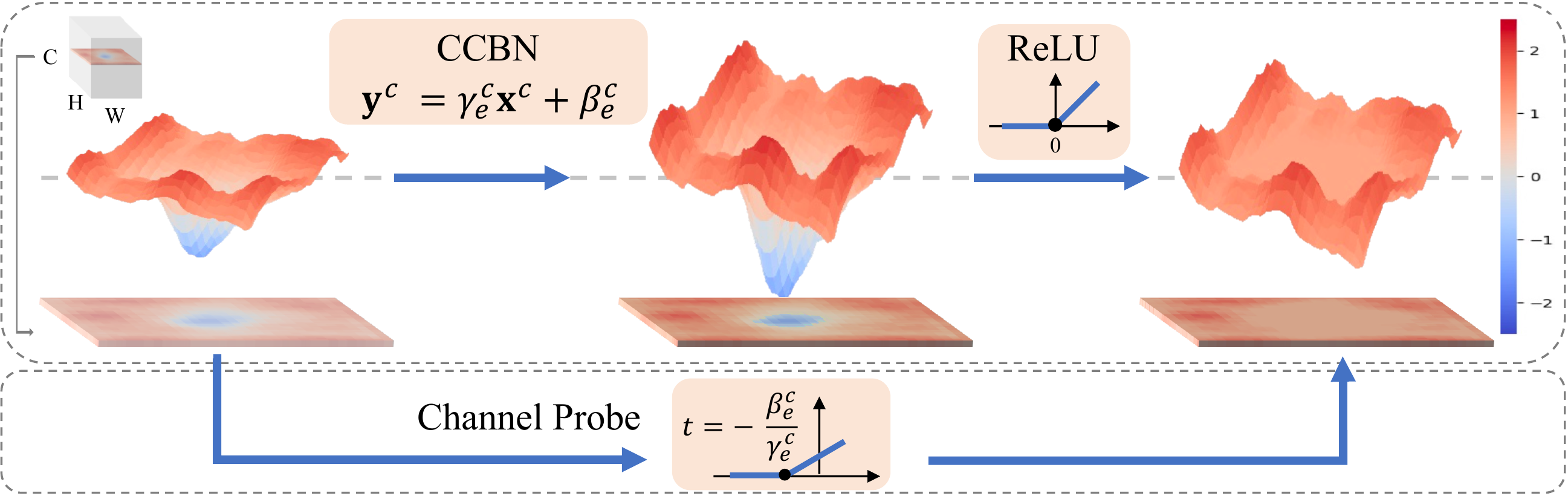}
    \vspace{-5pt}
    \caption{
        \textbf{Concept diagram of channel probe}, from whose statistics we develop the \textit{channel awareness}.
        For a particular feature channel, it is first transformed by class-conditional batch normalization (CCBN) with category-specific scale $\gamma_e^c$, and bias $\beta_e^c$, then activated by ReLU to filter out negative neurons.
        Such a process is reformulated as a combined operation (as shown at the bottom), where we define the channel probe as $t = - \frac{\beta_e^c}{\gamma_e^c}$.
        This value reflects the integrated action of CCBN and ReLU.
    }
    \label{fig:framework}
    \vspace{-10pt}
\end{figure*}

\subsection{Related Work}\label{sec:related-work}

Among various types of generative models, such as variational auto-encoder (VAEs)~\cite{kingma2013auto, higgins2016beta, van2017neural, razavi2019generating}, flow-based model~\cite{rezende2015variational, kingma2018glow}, diffusion model~\cite{ho2020denoising, dhariwal2021diffusion}, \textit{etc.}, GAN~\cite{gan} has received wide attention due to its impressive performance on both unconditional synthesis~\cite{pggan, stylegan, stylegan2, stylegan3} and conditional synthesis~\cite{sagan, sngan, biggan, stylegan-xl}.
Early studies on interpreting GANs~\cite{bau2019gandissect, yang2019semantic, shen2020interfacegan, editinginstyle} suggest that, a well-learned GAN generator has encoded rich knowledge that can be promising applied to various downstream tasks, including attribute editing~\cite{bau2020rewriting, gansteerability, ling2021editgan, bau2021paint, zhu2022region, zhang2021decorating, yang2019semantic}, image processing~\cite{gu2020image, pan2020dgp, huh2020ganprojection, zhu2020indomain, richardson2021encoding}, super-resolution~\cite{menon2020pulse, chan2021glean}, image classification~\cite{xu2021generative}, semantic segmentation~\cite{zhang2021datasetgan, 2021linearsemantic, tritrong2021repurposing, abdal2021labels4free, li2022bigdatasetgan}, and visual alignment~\cite{peebles2021gan}.
Existing interpretation approaches usually focus on the relationship between the latent space and the image space~\cite{shen2020interfacegan, zhu2021lowrankgan, yang2019semantic, wu2020stylespace}, hence commonly evaluated on unconditional models.
Some attempts are made to also analyze class conditional models~\cite{gansteerability, plumerault2020controlling, voynov2020unsupervised, ganspace, shen2021closed}, but they still target the latent space, leaving it unclear how the generator leverages the categorical information.
This work clearly \textbf{differs} from prior arts from the following aspects.
(1) We inspect the conditional generator from the channel perspective, which aggregates the messages from \textit{both the latent code and the class embedding}. To our knowledge, this is the first attempt on understanding the function of embedding space in conditional generation.
(2) We demonstrate the editability of altering a \textit{single channel} of the conditional generator. Different from the single-channel editing in unconditional GANs~\cite{wu2020stylespace}, our approach identifies different relevant channels with respect to different categories in an \textit{unsupervised} manner.
(3) We achieve fine-grained semantic segmentation by paying more attention to some ``important'' channels. Unlike the existing efforts on single-object generation~\cite{zhang2021datasetgan, 2021linearsemantic}, our method does \textit{not} require data-driven learning and can be robustly generalized to all classes.
(4) We also enable some applications that are \textit{peculiar to conditional models}, including the category hybridization and the category-wise synthesis performance evaluation.

\section{Methodology}\label{sec:method}

In this section, we introduce the proposed channel awareness.
Specifically, we re-examine the class-conditional batch normalization (CCBN), which is widely used in class conditional generation~\cite{biggan, sagan, sngan, contragan, projgan}, and investigate how it helps the generator with the categorical information provided by the class embedding.
It is noteworthy that our approach is \textit{fully unsupervised}, without relying on any segmentation masks or annotations.

\subsection{Preliminaries}\label{subsec:preliminary}

Unlike unconditional GANs~\cite{gan, pggan, stylegan} where the generator takes the latent code $\z$, as the only input, a class conditional generator employs an additional embedding vector, $\e$, to provide the categorical information.
Accordingly, the generation process can be formulated as $\I = G(\z, \e)$, where $\I$ and $G(\cdot, \cdot)$ are the output image and the generator, respectively.
That way, given a different embedding, the generator is able to produce images for that category.

There are many ways of integrating $\e$ into $G(\cdot, \cdot)$~\cite{stylegan, biggan}, where the most popular one is to adopt the class-conditional batch normalization (CCBN)~\cite{biggan, sagan, sngan, contragan, projgan}.
In particular, CCBN learns class-specific parameters to scale and shift the input feature maps, as
\begin{align}
    \y = {\bm\gamma}(\mathtt{concat}(\z, \e)) \odot \frac{\x - \mu(\x)}{\sigma(\x)} + {\bm\beta}(\mathtt{concat}(\z, \e)),  \label{eq:ccbn}
\end{align}
where $\x$ and $\y$, both with shape $C \times H \times W$, denote the input and output features.
$\mu(\cdot)$ and $\sigma(\cdot)$ compute the mean and variance of a tensor along the spatial dimensions (\textit{i.e.}, $H$ and $W$).
$\mathtt{concat}(\cdot, \cdot)$ stands for the concatenation operation.
${\bm\gamma}(\cdot)$ and ${\bm\beta}(\cdot)$ outputs the $C$-dimensional scale and bias by learning from both $\z$ and $\e$.
$\odot$ represents the element-wise multiplication with broadcasting.

\begin{figure*}[t]
    \centering
    \includegraphics[width=0.95\linewidth]{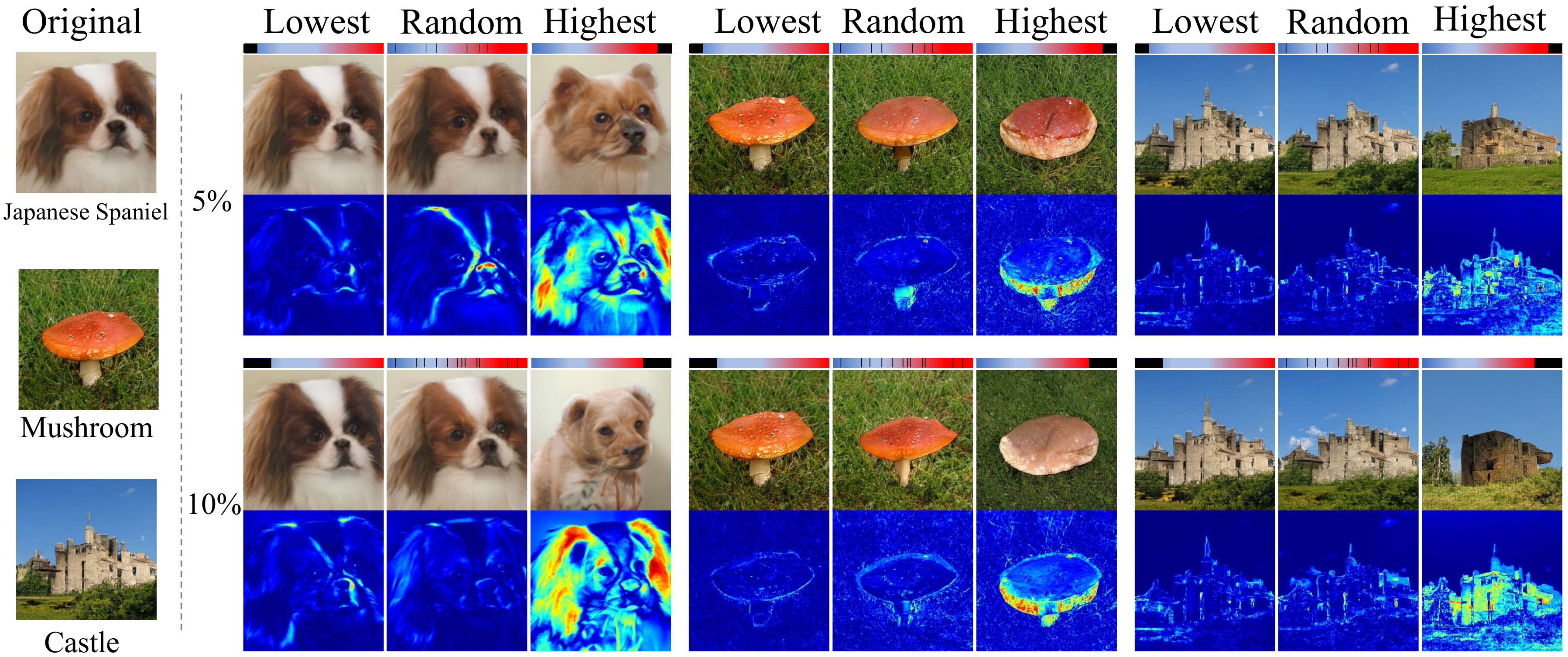}
    \vspace{-5pt}
    \caption{
        \textbf{Qualitative evaluation of category-oriented channel awareness} by \textit{zeroing out} a proportion of feature channels.
        \textit{Left:} Original synthesis.
        \textit{Right:} Distorted results and difference maps.
        Channels with the lowest, random, and highest awareness are selected.
        It verifies the effectiveness of the proposed channel awareness in measuring how a feature channel is relevant to synthesizing a category.
    }
    \label{fig:class_channels_quali}
    \vspace{-10pt}
\end{figure*}

\subsection{Channel Awareness}\label{subsec:channel_awareness}

\noindent\textbf{Channel Probe.}
From \cref{eq:ccbn}, we can tell that both the latent code and the class embedding act on the generation through CCBN.
In other words, their messages are delivered to the feature channels through the learning of ${\bm\gamma}(\cdot)$ and ${\bm\beta}(\cdot)$.
Now, we take a look at how each single channel contributes to the synthesis.
For a particular channel with index $c$, \cref{eq:ccbn} can be simplified as
\begin{align}
    \y^c = \gamma_e^c \x^c + \beta_e^c,  \label{eq:ccbn_channel}
\end{align}
where $\x^c$ and $\y^c$, both with shape $1 \times H \times W$, denote the normalized input (\textit{i.e.}, subtracting mean and dividing by standard deviation) and the output.
$\gamma_e^c$ and $\beta_e^c$ are scalars for the $c$-th channel.
Here, $e$ stands for the embedding index, carrying the categorical information, and the effect of $\z$ is omitted for simplicity.

CCBN is usually followed by ReLU~\cite{sagan, biggan}.
Such an activation controls the information flow in that negative values in $\y^c$ are cut off to zero.
According to \cref{eq:ccbn_channel}, it is equivalent to cutting off the values in $\x^c$ that are smaller than
\begin{align}
    \t = - \frac{\beta_e^c}{\gamma_e^c}.  \label{eq:probe}
\end{align}
In this way, we manage to directly relate the output feature channel to its corresponding input channel, as shown in \cref{fig:framework}.
The value $\t$ acts like a \textit{channel probe} since it measures the channel-wise activation of a particular synthesis.
Generally, for a certain channel, it presents different behaviours (\textit{i.e.}, with different $\t$ values) for different samples and different categories.

\noindent\textbf{Category-oriented Channel Awareness.}
As discussed above, given a well-trained generator with CCBN, $\t$ is strictly determined by the latent code, $\z$, and the class embedding, $\e$.
Recall that the channel probe $\t$ is instance-aware, which fluctuates along with the synthesis varying.
To get a more reliable understanding of the function of a single feature channel, we derive \textit{channel awareness} from the statistics of $\t$.
Oriented to the class embedding, $\e$, we would like to eliminate the impacts caused by the randomness of $\z$.
To this end, we first sample a number of latent codes with the embedding fixed, then calculate the $\t$ value for each synthesis and perform averaging.
Recall that a lower $\t$ suggests that more information will be preserved along this channel.
Therefore, channels with lower mean values are more likely to be used for generating the target class.
We define category-oriented channel awareness as $-\mathtt{E}_z[\t]$.

\noindent\textbf{Latent-oriented Channel Awareness.}
We further study the contribution of the latent code, $\z$, to each channel.
Similarly, we sample a collection of latent codes with a fixed embedding and calculate $\t$.
Differently, this time we are interested in the variance instead of the mean, which gives us the latent-oriented channel awareness.
A larger variance indicates that the randomness has a stronger influence on the synthesis regarding this channel.
Hence, the latent-oriented channel awareness, $\mathtt{Var}(\t)$, reflects how sensitively a channel reacts to the latent code with the given embedding.

\section{Evaluation and Analysis}\label{sec:analysis}
In this section, we evaluate and analyze the proposed channel awareness of category and latent in \cref{subsec:class_channels_eval} and \cref{subsec:latent_channels_eval}, respectively.
We exploit the BigGAN \cite{biggan} conditionally trained on the large-scale ImageNet \cite{imagenet} as the target model, which can generate realistic images of 1,000 various categories at 256 $\times$ 256 resolution . 
%

\subsection{Category-oriented Channel Awareness}\label{subsec:class_channels_eval}

\noindent\textbf{Qualitative Evaluation.}
We measure the causal channel effect on the image generation via forcing the target channel to zero during the forward process~\cite{bau2019gandissect}.
Then we compare the results after zeroing out of three groups of channels with different awareness scores in \cref{fig:class_channels_quali}.

We noticeably observe that: Zeroing out channels with the highest class awareness removes essential information of that class (\textit{e.g.}, fur type of Japanese Spaniel, stem of mushroom, and structure of castle), while masking channels with the lowest class awareness only causes negligible changes, and randomly selected channels have the in-between effect.
This phenomenon not only shows the effectiveness of our method but also demonstrates our \textit{key finding} that channels are not equally contributing to the generation of a particular class:
Only partial channels (\textit{i.e.}, with highest class awareness) primarily contribute to it, and some channels barely affect its generation (\textit{i.e.}, with lowest class awareness).

\begin{figure*}[t]
    \centering
    \includegraphics[width=1.0\linewidth]{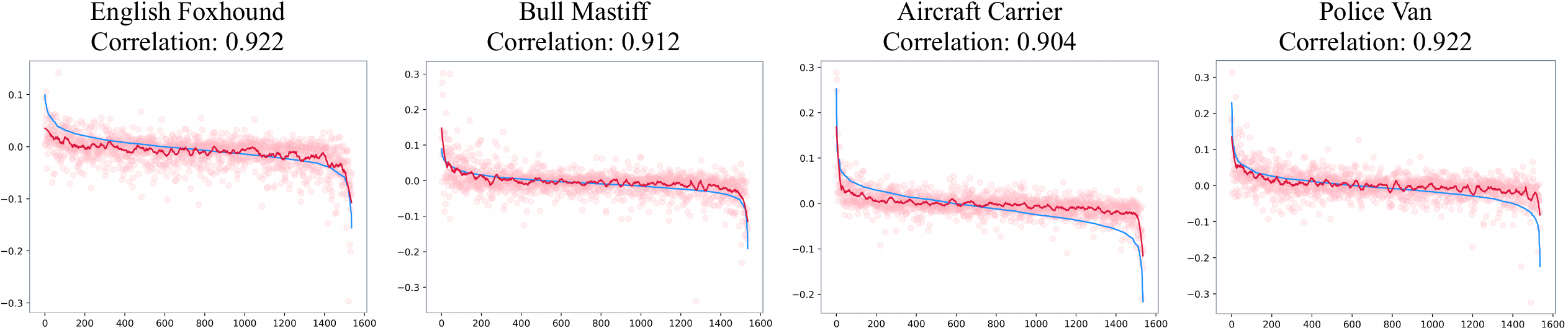}
    \vspace{-18pt}
    \caption{
        \textbf{Quantitative evaluation of category-oriented channel awareness} with \textit{single-channel modulation}.
        Each \textcolor{blue}{blue} curve exhibits the awareness score, while each \textcolor{red}{red} curve (smoothed with a Savitzky-Golay filter) presents the categorical score change measured by a pre-trained classifier.
        Their correlation appears on top of each figure.
        Note that the channel indices are sorted by the awareness scores.
    }
    \label{fig:class_channels_quati}
    \vspace{-10pt}
\end{figure*}

\noindent\textbf{Quantitative Evaluation.} 
We further quantitatively verify the \classaware regarding every single channel in all 1,000 classes.
We utilize a pre-trained classifier as an channel effect evaluator by assessing generated class images with and without a certain channel.
Specifically, we measure the classification score drop between generated image with zeroing out a channel and the original one, which we refer to as channel-classifier response.
We report the correlation coefficient between the channel-classifier response and the proposed \classaware.
High correlation indicates the effectiveness of our method.

\textit{Implementation details.} 
We use the Inception-v3~\cite{szegedy2016rethinking} model trained on ImageNet~\cite{imagenet} as the evaluation classifier, a widely used classifier for assessing generated images~\cite{heusel2017gans, impr}.
For each channel, we calculate its channel-classifier response among 1,000 generated sample pairs (a sample pair means the generated images before and after the intervention of single-channel masking).
To reduce data noise, we smooth the raw data of channel-classifier response via the commonly-used Savitzky-Golay filter~\cite{sgfilter} with a window length 51 (It is a tiny length of the filter since the full number of data points is 1,536).

We show four classes with the highest correlation in \cref{fig:class_channels_quati}.
The trend of channel-classifier response is highly aligned with that of \classaware.
We also calculate the average correlation for all 1,000 classes, even including categories that BigGAN does not synthesize well, and obtain 0.503.
Furthermore, we evaluate our method on BigGAN-deep\cite{biggan} regarding 1,000 classes, and obtain an average correlation of 0.603, even better than our base model.
Details can be found in Appendix~\ref{appendix:sec:deep}.
This result indicates the effectiveness of our metric regarding all classes as well as different models.

\begin{figure*}[t]
    \centering
    \includegraphics[width=0.95\linewidth]{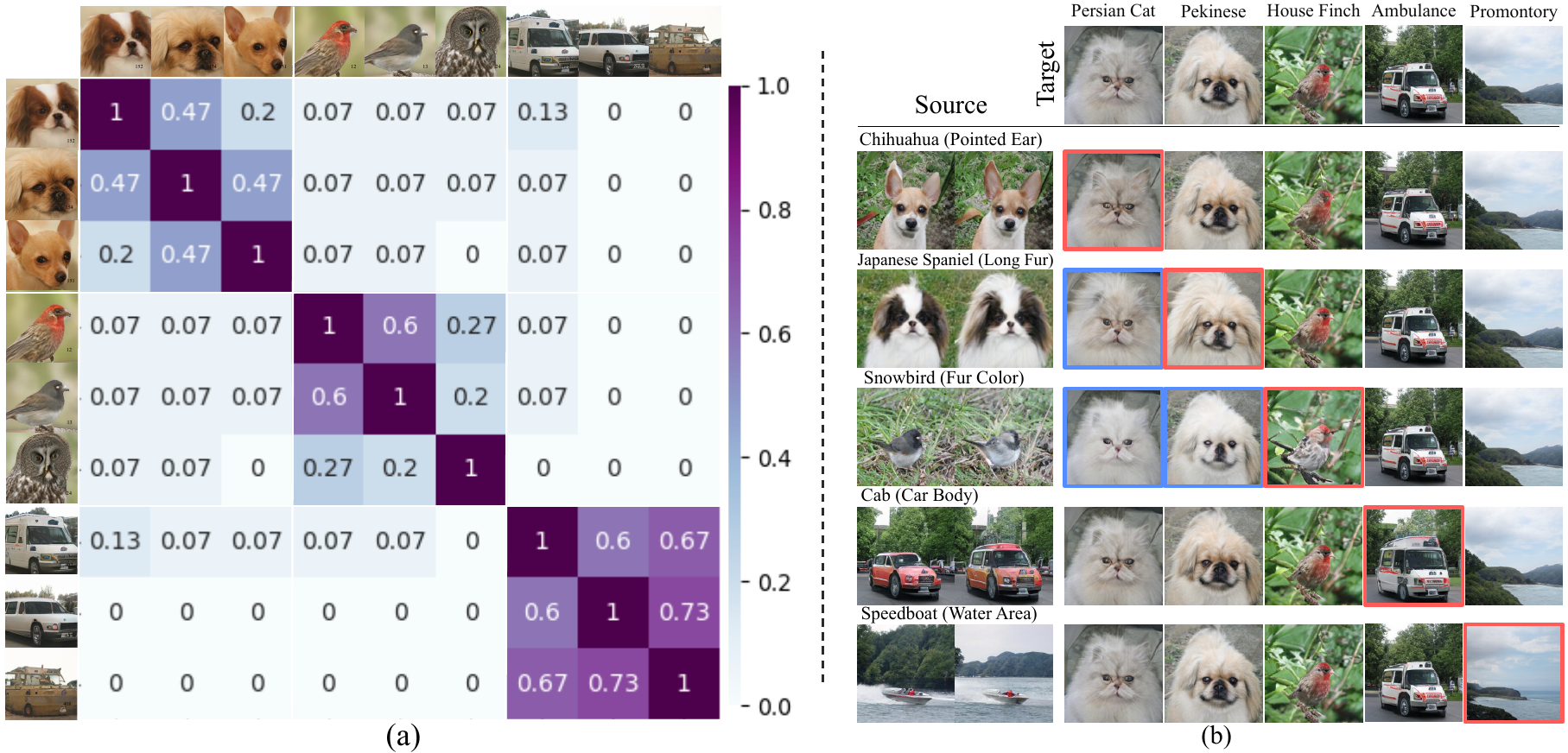}
    \vspace{-5pt}
    \caption{
        \textbf{Analysis on category-oriented channel awareness.}
        (a) Channel overlap between different categories (including dogs, birds, and vehicles), where a higher overlap indicates two classes sharing more relevant channels.
        The overlap is computed using 100 channels with the highest awareness for each category.
        (b) The semantic discovered from the source class is applied to other target classes.
        We observe that semantics are shared between similar categories, like cat and dog.
    }
    \label{fig:cross_class_analysis}
    \vspace{-5pt}
\end{figure*}

\noindent\textbf{Analysis via channel overlap estimation.}
From above, we verify that channels with high awareness of a given class are primarily responsible for generating that particular category.
Next, we wonder how distinct those channels are regarding different categories, 
and do similar categories share several channels?
Answering this question could help us deeply understand how a conditional GAN organizes characteristics from numerous distinct categories.
Thus, we further analyze the channel overlap among them.
We estimate the channel overlap of two classes as $ \frac{ \vert S_i \cap S_j \vert } {k} $,
where $ S_i $ and $ S_j $ represent the set of channel indices with top k highest awareness of class $ i $ and $ j $,
$ \cap $ denotes the intersection operation, and $ k  =|S_i|=|S_j|$, which is the number of selected channels. 
For the given two classes, higher channel overlap indicates that more channels are responsible for generating both of them, and zero channel overlap means they own two separate sets of channels inside the network.

The results of channel overlap of different categories are shown in \cref{fig:cross_class_analysis}a.
We can observe that:
1) Channels with high class awareness are different between different classes since their overlaps are less than $1$.
2) Several channels are overlapped among similar classes (see the three $ 3 \times 3 $ blocks along the diagonal).
3) For two largely distinct classes, such as cars with dogs and birds, their channel overlaps are close to zero, meaning that their corresponding channels are almost disjoint.
These observations indicate that different classes own different sets of relevant channels, and similar classes own several mutual relevant channels.

\noindent\textbf{Analysis via single-channel manipulation.}
Furthermore, we discover that altering a single channel of the highest awareness can manipulate the class-relevant semantics which results are shown in \cref{fig:cross_class_analysis}b (left two columns), such as the pointed ear of class Chihuahua and the long fur of class Japanese Spaniel.
Meanwhile, when applying the semantic found by one source class to other target classes, we found that this can only manipulate similar classes (\textcolor{red}{red} box), or other classes that exhibit the corresponding semantic (\textcolor{blue}{blue} box), while does not affect other irrelevant classes.
For example, altering the semantic of water area (last row) found by the class of speedboat can manipulate the promontory class yet not affect animal classes, same as other semantics.

\setlength{\tabcolsep}{8pt}
\begin{table*}[t]
    \centering
    \caption{
        \textbf{Cross-category analysis.}
        Regarding the 1,000 classes in ImageNet~\cite{imagenet}, we select only 10 channels for each class according to different criteria, and find the common channels through intersection.
        Results on all layers suggest that our \textit{latent-oriented channel awareness} helps identify channels that can deliver information to all classes.
    }
    \label{tab:CO_latent_channels}
    \vspace{-8pt}
    \begin{tabular}{lcccccccccccc}
        \hline
        Criterion & L0 & L1 & L2 & L3 & L4 & L5 & L6 & L7 & L8 & L9 & L10 & L11 \\
        \hline
        highest awareness \textit{w.r.t.} category & 0 & 0 & 0 & 0 & 0 & 0 & 0 & 0 & 0 & 0 & 0 & 0 \\
        lowest awareness \textit{w.r.t.} category  & 0 & 0 & 0 & 0 & 0 & 0 & 0 & 0 & 0 & 0 & 0 & 0 \\
        highest awareness \textit{w.r.t.} latent   & 3 & 3 & 3 & 1 & 3 & 3 & 0 & 4 & 2 & 1 & 5 & 4 \\
        \hline
    \end{tabular}
    \vspace{-5pt}
\end{table*}

\subsection{Latent-oriented Channel Awareness}\label{subsec:latent_channels_eval}

\begin{figure*}[t]
    \centering
    \includegraphics[width=0.95\linewidth]{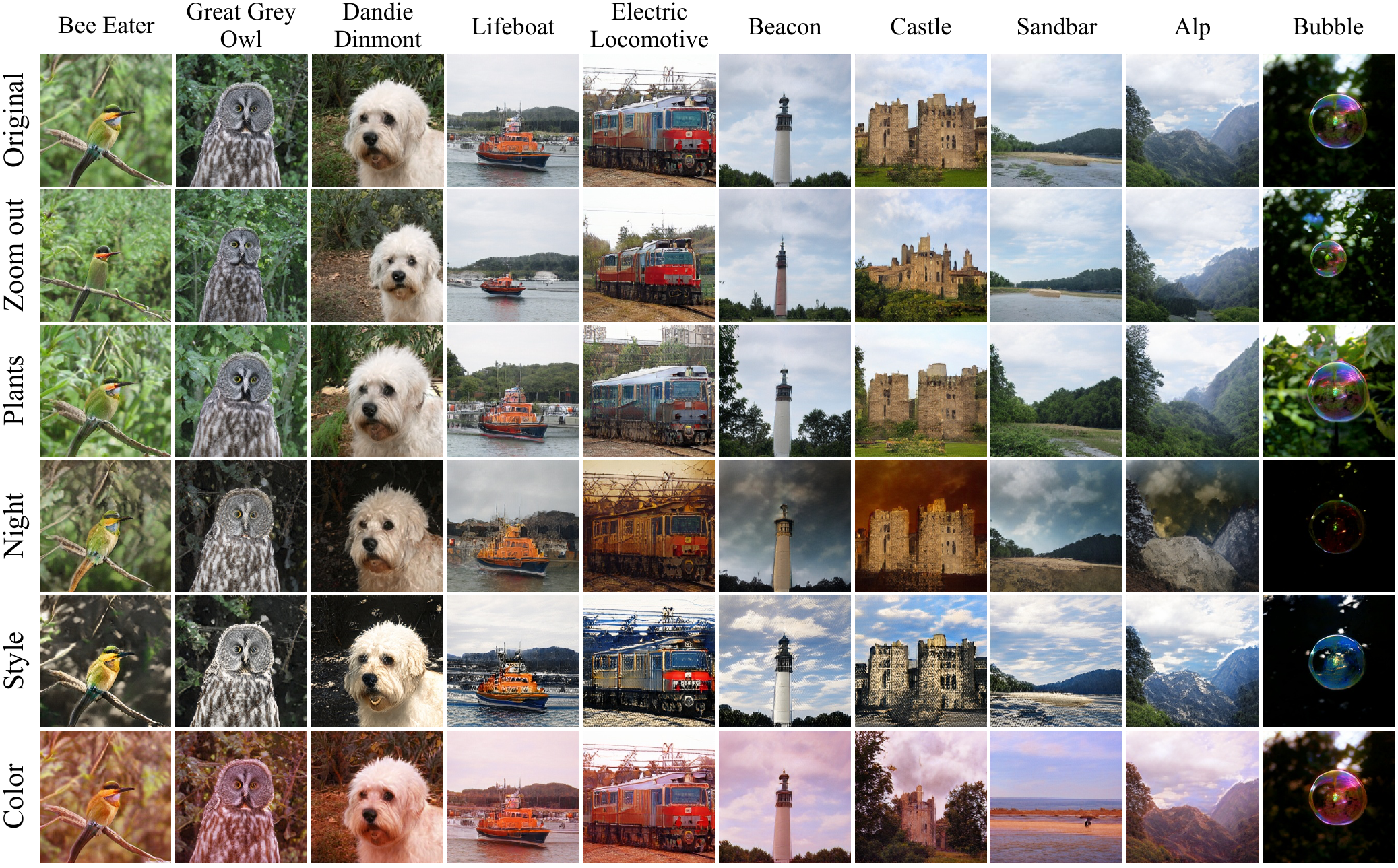}
    \vspace{-5pt}
    \caption{
        \textbf{Versatile latent-oriented attribute editing results} achieved by \textit{single-channel} modulation.
        We manage to find the channels that are shared by all classes.
        The details of identifying these channels are described in \cref{subsec:latent_channels_eval} and \cref{tab:CO_latent_channels}.
    }
    \label{fig:class_agnostic_attribute_editing}
    \vspace{-10pt}
\end{figure*}

\noindent\textbf{Discovery of class-shared channels.}
From above, we show that channels with high \classaware are class-distinct thus responsible for class-distinct semantics.
We also aim to inspect if there exist channels that took charge of common characteristics for all 1,000 classes, which we refer to as class-shared channels.
To answer this question, we perform channel intersection among all classes to check whether there has left channels.
Specifically, we select class-specific channels of 1) highest awareness to the class, 2) lowest awareness to the class, and 3) highest awareness to latent, and then perform intersection successively.
The channel intersection results of these three settings are shown in \cref{tab:CO_latent_channels}.
We can discover that after intersecting among 1,000 classes, there is no channel left in the first two settings.
However, we noticeably identified that channels that highly respond to latent exist in all 1,000 classes.
This implies that a class-conditional generator can learn unified representations provided to all classes.
Besides, these class-shared channels can further enable semantic manipulation with highly similar effect on disparate classes.
Visual editing results with respect to these channels are provided in \cref{fig:class_agnostic_attribute_editing}.

\section{Applications with Class Conditional GANs}\label{sec:application}

In this section, we demonstrate four novel applications enabled by our channel awareness.
(1) Versatile image editing for category-oriented attributes and latent-oriented attributes (\cref{subsec:single_channel_manipulation}).
(2) Category hybridization that smoothly blends attributes from two different categories (\cref{subsec:channel_mixing}).
(3) Fine-grained Segmentation on the synthesized image (\cref{subsec:segmentation}).
(4) Category-wise synthesis evaluation from the perspective of internal parameters of GANs (\cref{subsec:total_channel_awareness}).

\begin{figure*}[t]
    \centering
    \includegraphics[width=0.95\linewidth]{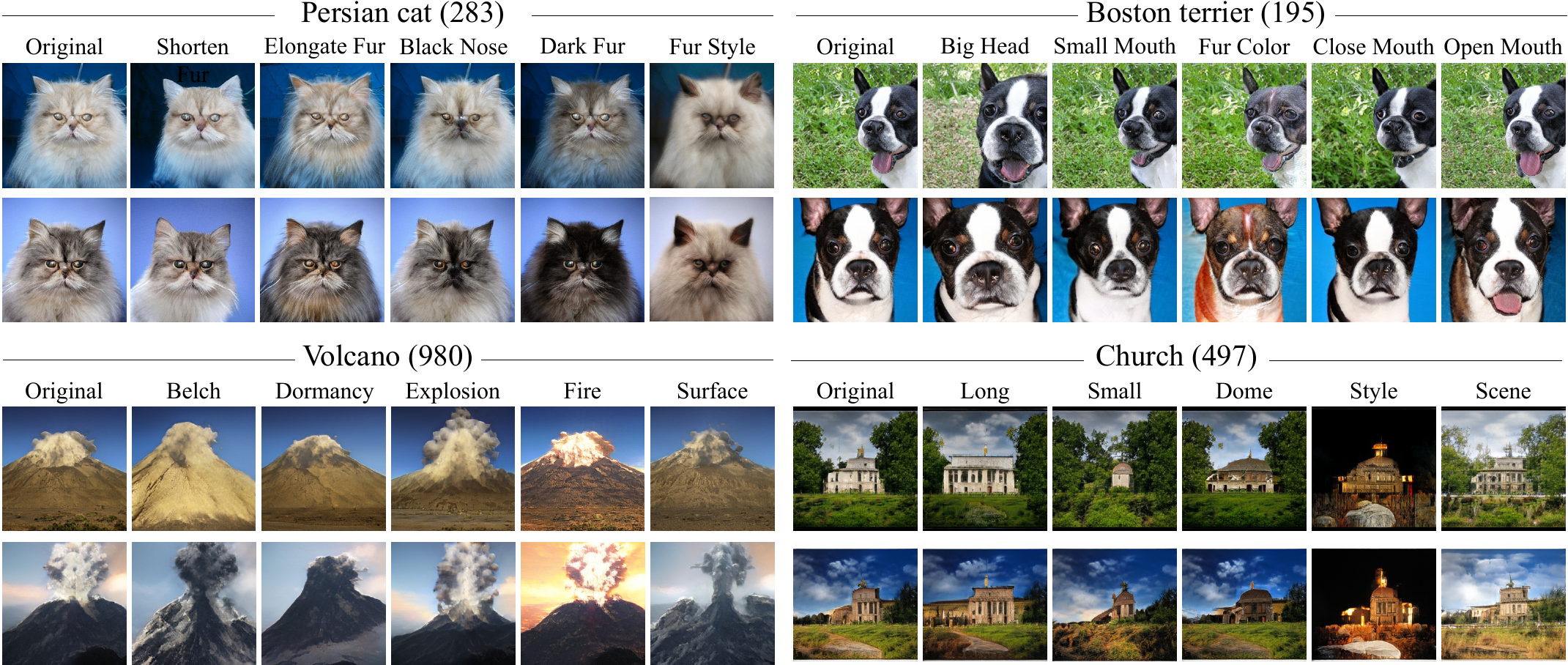}
    \vspace{-5pt}
    \caption{
        \textbf{Versatile category-oriented attribute editing results} achieved by \textit{single-channel} modulation.
        We manage to find the channels most relevant to a particular class by selecting those with high category-oriented channel awareness scores.
    }
    \label{fig:class_relevant_attribute_editing}
    \vspace{-5pt}
\end{figure*}

\begin{figure*}[t]
    \centering
    \includegraphics[width=1.0\linewidth]{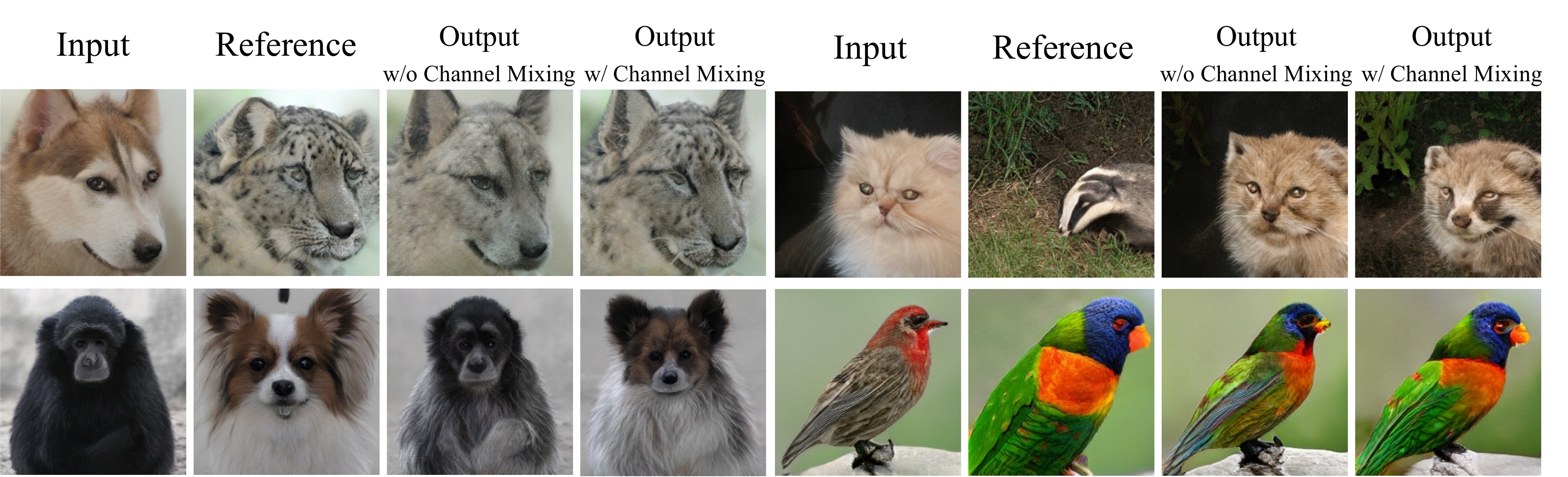}
    \vspace{-20pt}
    \caption{
        \textbf{Category hybridization} by mixing the channels that are relevant to two different categories.
        Different from the conventional style mixing~\cite{stylegan} (\textit{i.e.}, the third column), our approach (\textit{i.e.}, the fourth column) better fuses the characteristics (including both shape and appearance) of both classes.
    }
    \label{fig:channel_mixing}
    \vspace{-10pt}
\end{figure*}

\subsection{Image Editing via Single-channel Manipulation}\label{subsec:single_channel_manipulation}

With our method, we can identify meaningful channels through awareness ranking, from which we discover that simply manipulate a single channel can achieve versatile effects for image attribute editing.
Such a manipulation can be either multiplying or adding a constant of manipulation magnitude.
Therefore, with channels that are relevant to a particular class, we can discover and then manipulate various attributes of this class, such as fur, mouth, and nose for dogs and cats, explosion for volcanic. 
Results of image editing about class-oriented attribute are shown in \cref{fig:class_relevant_attribute_editing}.
Meanwhile, with channels that respond to latent, we can edit meaningful class-shared semantics controlled by the latent vector, which can therefore manipulate all kinds of classes in BigGAN.
Results of image editing about latent-oriented attributes are shown in \cref{fig:class_agnostic_attribute_editing}.

\textit{Discussions and comparisons.}
Compared with prior works which present manipulation results on generated images of BigGAN via manipulating attribute vectors in the latent space~\cite{voynov2020unsupervised, gansteerability, shen2021closed, ganspace}, our method demonstrates that simply altering only one feature channel could achieve this.
Besides, our method can retrieve semantics controlled by both latents and class embeddings, rather than latent only.
Compared with StyleSpace~\cite{wu2020stylespace}, and GAN dissection~\cite{bau2019gandissect}, which edit on a single channel, our method has two obvious differences: 
1) Our method does not require any semantic segmentation masks and data collection for annotation. 
Instead, we discover channels in a fully unsupervised manner. 
2) Our method targets class conditional GAN, which can synthesize 1,000 categories rather than unconditional GANs, limited to single class modeling.

\subsection{Category Hybridization via Channel Mixing}\label{subsec:channel_mixing}
In this section, we define a novel image editing task called \textit{category hybridization}, which aims to synthesize realistic images while preserving meaningful attributes from more than one categories.
This task could benefit novel content creation and produce unnatural yet plausible combinations of two categories which do not exist in the training set.
%

Our method intrinsically fits the category hybridization task since we can locate class-distinct channel representations for different given classes.
After that, we transplant the representations of the reference class to the corresponding channel positions in the feature of the input class, which we refer to as \textit{channel mixing}.
To activate the transplanting channels, we exploit the class embedding of the reference class for a continued generation.
Results of category hybridization are shown in \cref{fig:channel_mixing}. 
We observed that with channel mixing, we could harmoniously fuse attributes from two categories into one realistic image which cannot achieve by style mixing (\textit{i.e.}, simply mixes layer-wise class embeddings).
These editing effects also validate the effectiveness of our channel awareness that we can discover class-oriented attributes.

\begin{figure*}[t]
    \centering
    \includegraphics[width=1.0\linewidth]{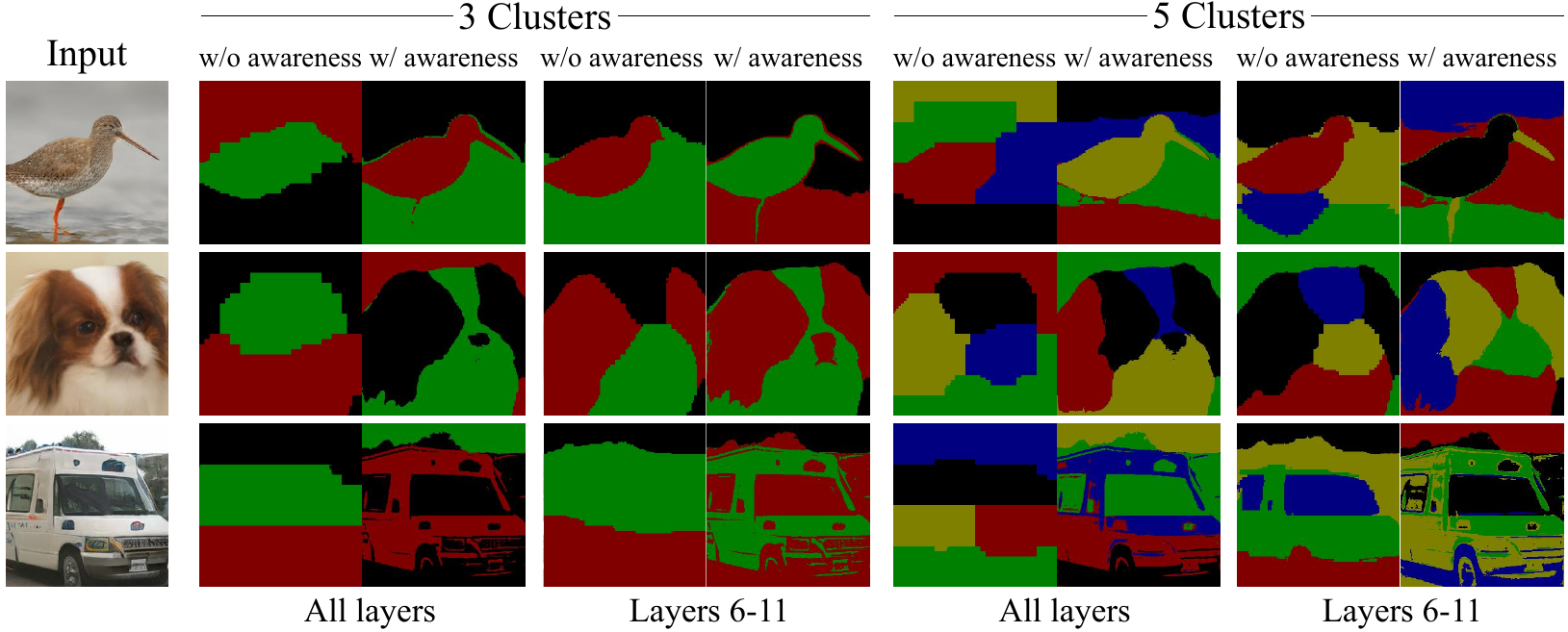}
    \vspace{-18pt}
    \caption{
        \textbf{Fine-grained semantic segmentation} with \textit{channel weighted clustering}.
        Concretely, we collect the multi-level features produced by the generator, weight them with the proposed channel awareness, and perform clustering on all pixels.
        It turns out that our approach helps the clustering to focus more on the relevant channels, and hence suggests a better performance.
    }
    \label{fig:segmentation}
    \vspace{-5pt}
\end{figure*}

\subsection{Segmentation via Channel Weighted Clustering}\label{subsec:segmentation}

Existing works~\cite{2021linearsemantic, zhang2021datasetgan, abdal2021labels4free, li2022bigdatasetgan, tritrong2021repurposing, ling2021editgan} revealed that internal features in GAN can enable object part segmentation of generated images, facilitating semantic annotation synthesis and local semantic editing.
Here, we present a simple prototype that utilizes \classaware to do unsupervised semantic segmentation on generated images.
Specifically, we collect the internal features from multiple ReLU layers of the generator, then upsample each feature to be the same size as the output image $H' \times W'$.
After upsampling, features from multiple layers could be concatenated along the channel dimension into a total feature volume as $C' \times H' \times W'$, where $C'$ is the total number of channels.

A pixel-wise feature vector in the volume, in size of $C' \times 1 \times 1$, is considered as a data sample that contains sufficient semantic information.
Then, we perform $K$-Means clustering on pixel-wise feature vectors to segment the image.

Recall that \classaware quantifies the importance of each channel to a target category.
Hence, we employ normalized channel awareness to weight those pixel-wise feature vectors before clustering.
Segmentation results without and with channel awareness are shown in \cref{fig:segmentation}.
We show the results of two feature collections (\textit{i.e.}, from all layers and the second half of layers) and two different numbers of clusters (\textit{i.e.}, three and five).
We can observe that our channel awareness can split very fine regions such as the beak and legs in the first image of the bird in \cref{fig:segmentation}.
This is due to our channel awareness could highlight important channel representations and therefore could obtain a more precise and finer segmentation map.

\begin{figure*}[t]
    \centering
    \includegraphics[width=1.0\linewidth]{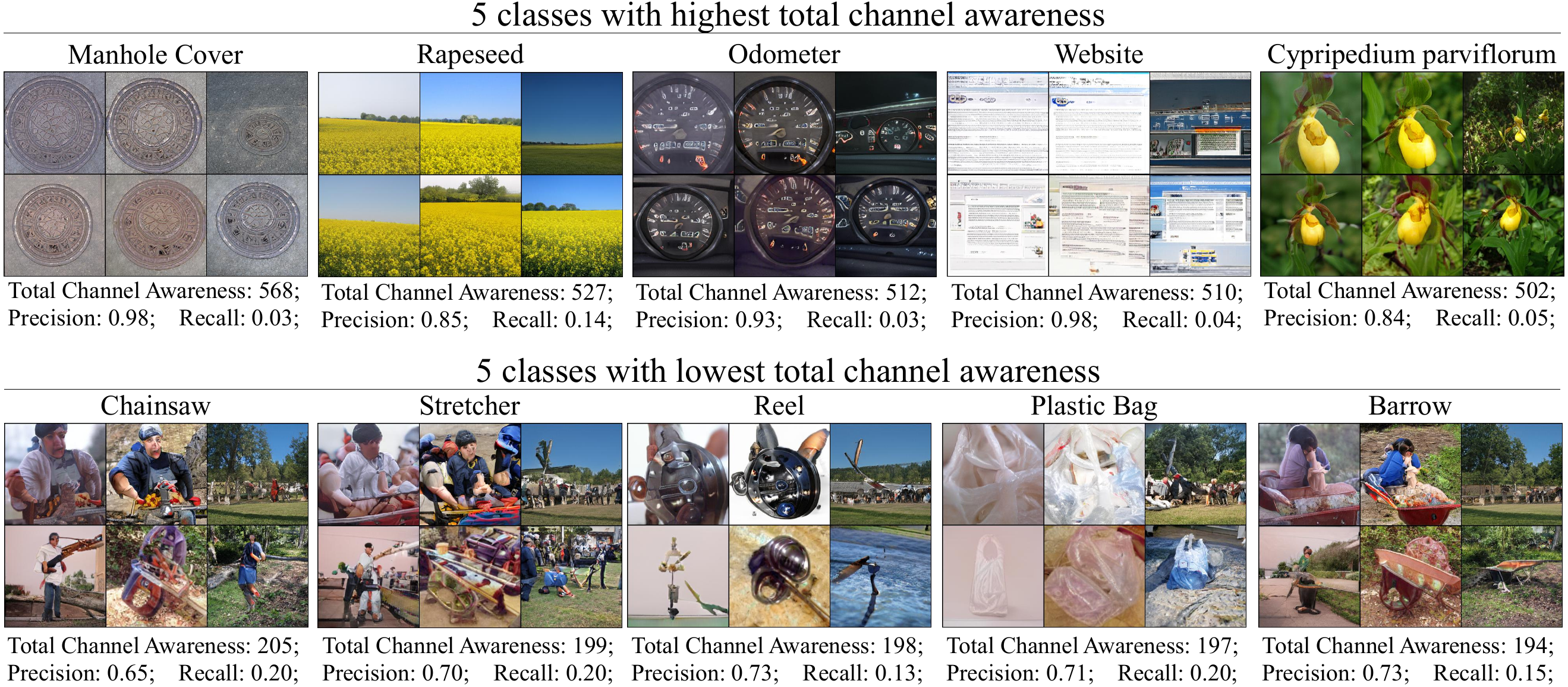}
    \vspace{-18pt}
    \caption{
        \textbf{Qualitative results of evaluating category-wise synthesis performance}, which is enabled by our \textit{total channel awareness}.
        We observe that classes with high total awareness tend to have high synthesis quality yet low diversity.
    }
    \label{fig:quality_interpretation}
    \vspace{-10pt}
\end{figure*}

\begin{figure*}[t]
    \centering
    \includegraphics[width=1.0\linewidth]{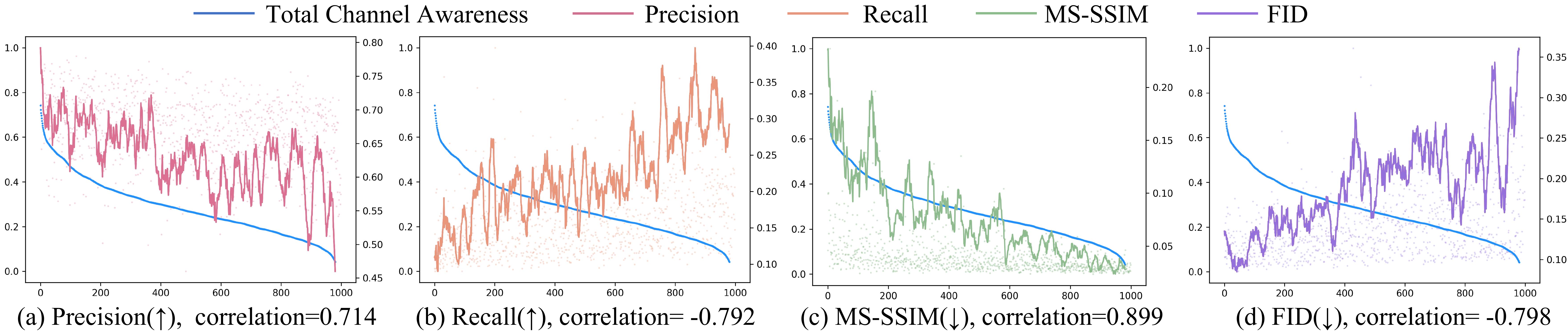}
    \vspace{-18pt}
    \caption{
        \textbf{Quantitative results of evaluating category-wise synthesis performance}, including the correlation between our \textit{total channel awareness} and the category-wise precision~\cite{impr}, recall~\cite{impr}, MS-SSIM~\cite{odena2017conditional}, and FID~\cite{heusel2017gans}.
        \textit{X-axis:} Class indices sorted by the total awareness score.
        \textit{Y-axis:} Normalized values for total channel awareness and other four metrics (left) and values for the smoothed curve (right).
        $\uparrow$ indicates higher value is better while $\downarrow$ indicates lower value is better.
        The correlations between total channel awareness and each candidate metric are shown at the bottom.
        Our total channel awareness is highly correlated with all of the above four metrics.
    }
    \label{fig:quality_interpretation_correlation}
    \vspace{-10pt}
\end{figure*}

\subsection{Synthesis Evaluation via Total Channel Awareness}\label{subsec:total_channel_awareness}
Reliable evaluation metrics can help researchers better assess and improve the overall performance of generative models.
In this section, we describe why and how our channel awareness is capable of evaluating the category-wise synthesis performance.

\textit{Assumption and method.} Our key assumption is that if one class has the \textit{awareness} of consistently activating certain channels for producing category-oriented attributes while deactivating certain channels for suppressing irrelevant attributes, 
then the category tend to provide faithful generation results. 
For example, generating a flower needs to suppress those channels responsible for eyes and nose (mainly used by other classes like animals) and activate those related to petals.
Thus, we measure the overall selection and suppression awareness as total channel awareness, by summing the absolute values of \classaware over all layers, which can be denoted as follows: $a_e=\sum_{c \in C'} \vert \mathtt{E}_z[\t] \vert$. 
Here $C'$ means the number of total channels among all layers and $e$ is the given class.

\textit{Results.} With the total channel awareness, we empirically find that classes with the higher total channel awareness exhibit better quality yet low diversity of generated samples. 
Qualitative results of generated samples of classes with the highest and lowest total channel awareness are shown in \cref{fig:quality_interpretation}.
For quantitative verification, we calculate the correlation between total channel awareness and other widely used metrics for GANs among all 1k classes.
Specifically, we estimate the quality via precision~\cite{impr}, diversity via recall~\cite{impr} and MS-SSIM~\cite{odena2017conditional}, and a general metric FID~\cite{heusel2017gans} for each class.
Implementation details can be found in Appendix~\ref{appendix:sec:eval_performance}.
We plot the results in \cref{fig:quality_interpretation_correlation}, with raw data (dots in light colors) smoothed by a sliding window of size 20 and a scaled y-axis for better view.
Note that for metrics taking diversity into major consideration (\textit{i.e.}, Recall and MS-SSIM), higher total channel awareness relates to lower diversity, which is aligned with the observation presented in \cref{fig:quality_interpretation}.

\textit{Discussions.}
Unlike FID, precision and recall, which are estimated via features extracted from a wide range of generated and real images (\textit{e.g.}, usually 50k generated images for FID), and MS-SSIM, which is calculated via similarity between sample pairs, our total channel awareness is calculated by the statistics of internal parameters in GANs (statistics of the proposed channel probe).
This difference makes our method own several obvious advantages as follows:
1) Our algorithm runs extremely faster than aforementioned metrics.
Instead of sampling numerous images and then extracting features via an extra classifier, channel awareness can be easily estimated via forwarding only the FC layer in the CCBN module.
2) Additionally, we provide a novel and complementary perspective for evaluating class conditional GANs based on the internal parameters inside a GAN. 
We provide insight that if a class has high channel awareness of activating and deactivating certain channels, then the class tends to generate well, based on which a better regularization for improving synthesis performance can be developed in future works.
3) Furthermore, as we illustrate that high-quality classes exhibit low intra-class diversity, this implies that the class generation tends to sacrifice one aspect to meet another, showing the quality-diversity trade-off the model exhibits.
This observation may provide future work with insights on how to better balance the this trade-off during learning.
\section{Conclusion and Discussion}\label{sec:conclusion}

This work takes the first step towards understanding the generation mechanism of conditional GANs from the channel perspective.
Concretely, we propose simple yet effective \textit{channel awareness}, by which we successfully identify that channels that response to the latent code and to different class embedding.
Extensive analyses shed light on how a conditional GAN manages the categorical information with different channels.
More importantly, our channel awareness enables four novel applications with class conditional generators, which are rarely explored by prior work.
In particular, we achieve single-channel attribute editing as well as harmonious category hybridization in a fully unsupervised manner.
We also demonstrate the promising potential of our channel awareness in fine-grained semantic segmentation and category-wise synthesis performance evaluation.

Despite the appealing results, there are still some directions worth exploring.
Recall that this work primarily targets the BigGAN generator~\cite{biggan}, which learns the categorical information through CCBN with ReLU activation.
Investigating other architectures, like the style modulation layer~\cite{stylegan} with Leaky ReLU activation~\cite{leakyrelu}, can be one of the future works.
Besides, our approach mainly focuses on interpreting and utilizing a well-learned model for downstream tasks.
How the insights provided in this work can inspire the design of a more powerful GAN model would be of more significance.

{\small
\bibliographystyle{abbrv}
\bibliography{ref}
}

\appendix
\section*{Appendix}
\begin{figure*}[t]
    \centering
    \includegraphics[width=1.0\linewidth]{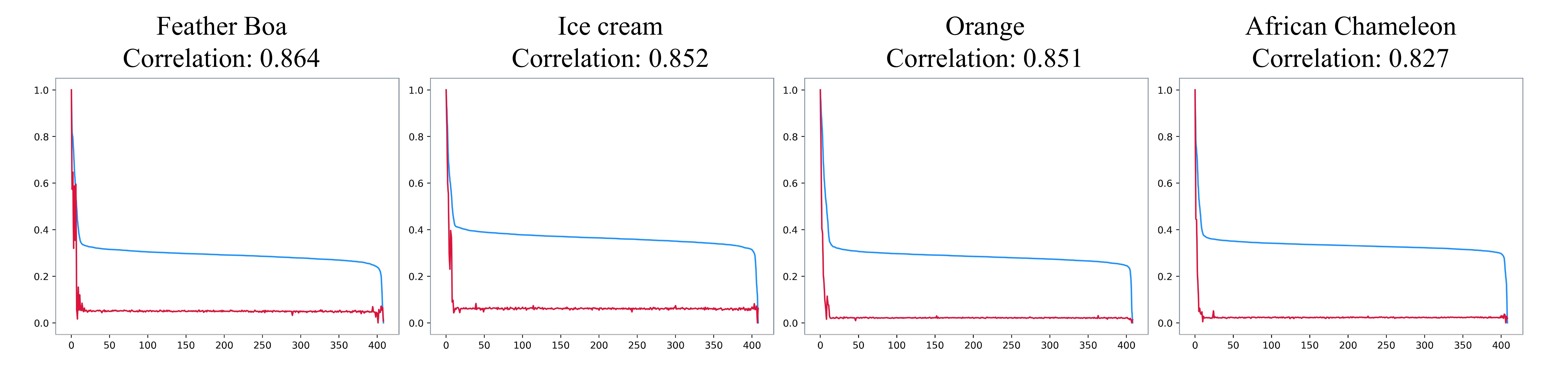}
    \vspace{-18pt}
    \caption{
        \textbf{Quantitative evaluation of category-oriented channel awareness on BigGAN-Deep} with five-channel modulation.
        The correlation between the awareness score (blue) and channel-classifier response (red) appears on top of each figure.
        \textit{X-axis:} intervention order following the sorted awareness score.
        \textit{Y-axis:} Scaled values of both lines.
        The average correlation of all 1,000 classes is 0.603.
    }
    \label{fig:deep_5chs}
    \vspace{-10pt}
\end{figure*}

\section{Evaluation on BigGAN-Deep}\label{appendix:sec:deep}

This section provides implementation details and results for evaluating the \classaware on the BigGAN-Deep model~\cite{biggan}.
Unlike BigGAN, BigGAN-Deep has twice the number of channels (\textit{e.g.}, 2,048 channels in layer 0), making the single-channel evaluation more time-consuming.
Besides, zeroing out a single channel makes an invisible change on the output image.
Thus in this experiment, we zero out five channels for each intervention, and every five channels are selected along with the order of sorted channel indices by the \classaware.
Results of top four results of four classes are shown in \cref{fig:deep_5chs}.
Since zeroing out five channels makes more changes than zeroing out one channel, here we need to scale the two curves into the range of 0-1 for a better view.
The average correlation of all 1,000 classes is 0.603.

\section{Synthesis Performance Evaluation}\label{appendix:sec:eval_performance}

This section gives the implementation details and results for category-wise synthesis performance evaluation via total channel awareness.

\noindent\textbf{Data preparation.} For precision, recall, and FID, which need to compare fake images with real images during estimation, we firstly pre-process real images from ImageNet~\cite{imagenet} following the procedure in BigGAN training which is cropping and resizing to 256 $\times$ 256.
The number of real images in ImageNet is around 1,300 for each class.
We randomly sample all generated images without truncation for fairly measuring the synthesis performance.

\noindent\textbf{Precision and Recall.}
Precision~\cite{impr} is used for measuring sample quality by retrieving fake images and checking whether it locates within the manifold of real images.
Recall~\cite{impr} is used to measure sample diversity by retrieving real images and checking whether it is within the manifold formed by fake images.
The ranges of these two metrics are both between zero to one.
A higher value indicates a better performance.
We calculate precision and recall with the number of fake images equal to the number of real images of each class followed the suggested setting~\cite{impr}.
The feature extraction network is the pre-trained VGG-16.
We set the neighborhood size $k$ equal to 3, which is a more robust choice\cite{impr}.

\noindent\textbf{MS-SSIM.}
We calculate the similarity between fake image pairs via the image similarity metrics MS-SSIM following~\cite{odena2017conditional}.
Lower image similarity indicates better diversity.
For each class, we randomly sample 100 fake images to construct 10,000 image pairs for calculating similarity, then an averaged similarity score for the class can be obtained as the final result for the intra-class diversity.

\noindent\textbf{Fréchet Inception Distance (FID).}
FID is a general metric that considers both diversity and quality by directly estimating the distance between the feature distribution of fake and real images.
We calculate the FID with 50,000 generated images for each class.
Then we exploit a pre-trained Inception-v3 as the feature extractor for estimating the feature distance between real and fake images.

\section{Efficiency of Computing Channel Awareness}\label{appendix:sec:code}

\begin{figure}[t]
    \centering
    \includegraphics[width=0.8\linewidth]{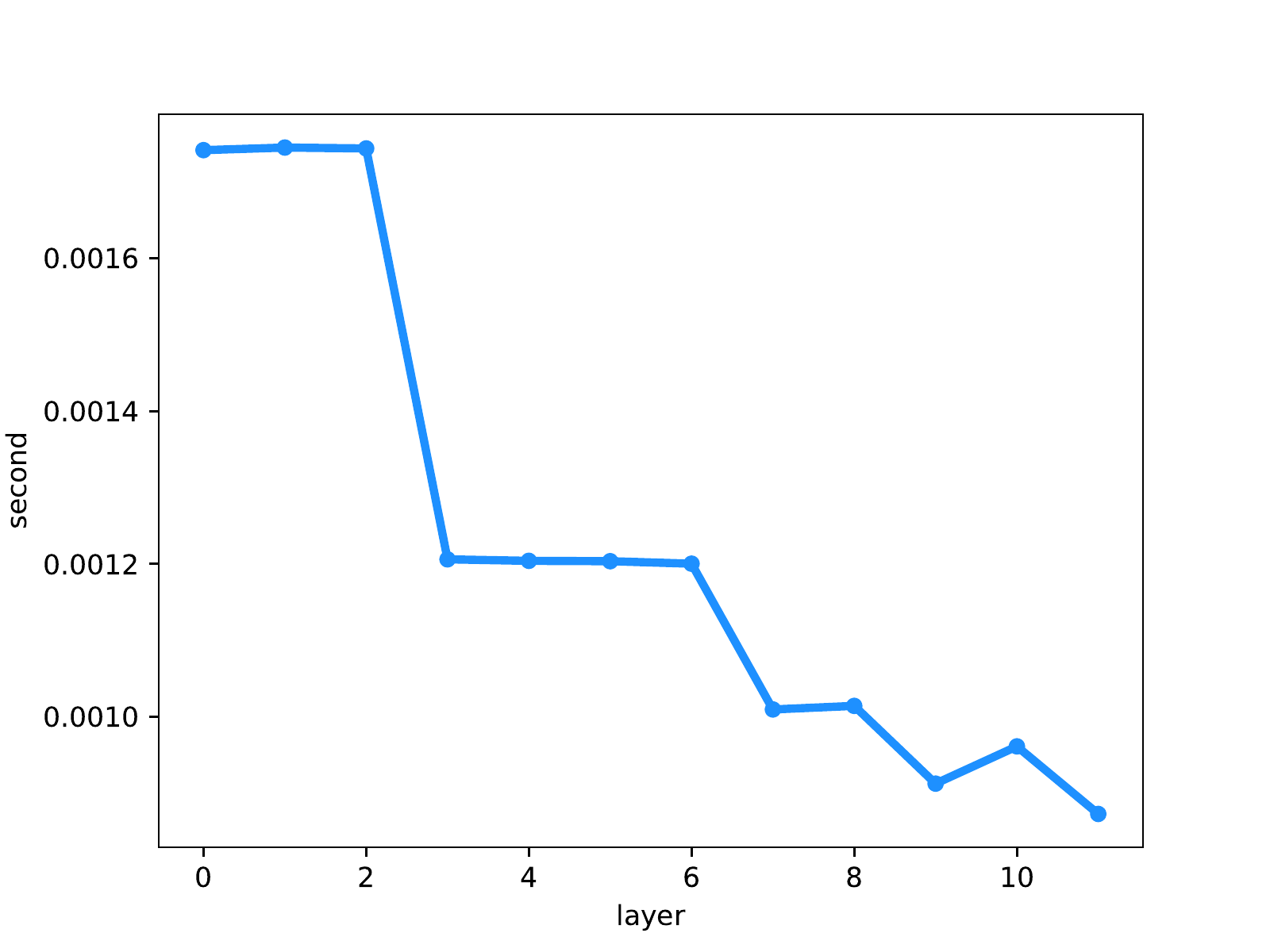}
    \vspace{-5pt}
    \caption{
        Layer-wise running time of interpreting BigGAN.
    }
    \label{fig:runtime}
    \vspace{-10pt}
\end{figure}

Given a pre-trained class conditional model, our proposed channel awareness, including both category-oriented channel awareness and latent-oriented channel awareness, can be computed \textit{unsupervisedly and efficiently}.
As shown in \cref{fig:runtime}, with one NVIDIA Tesla V100 GPU, it costs $0.0017s$ to interpret all channels in the first layer of BigGAN regarding a certain class.
The average running time for all layers is $0.0012s$.
Here, for each layer, the running time is averaged using 200 classes.

\section{More Results}\label{appendix:sec:results}

\begin{figure*}[t]
    \centering
    \includegraphics[width=0.82\linewidth]{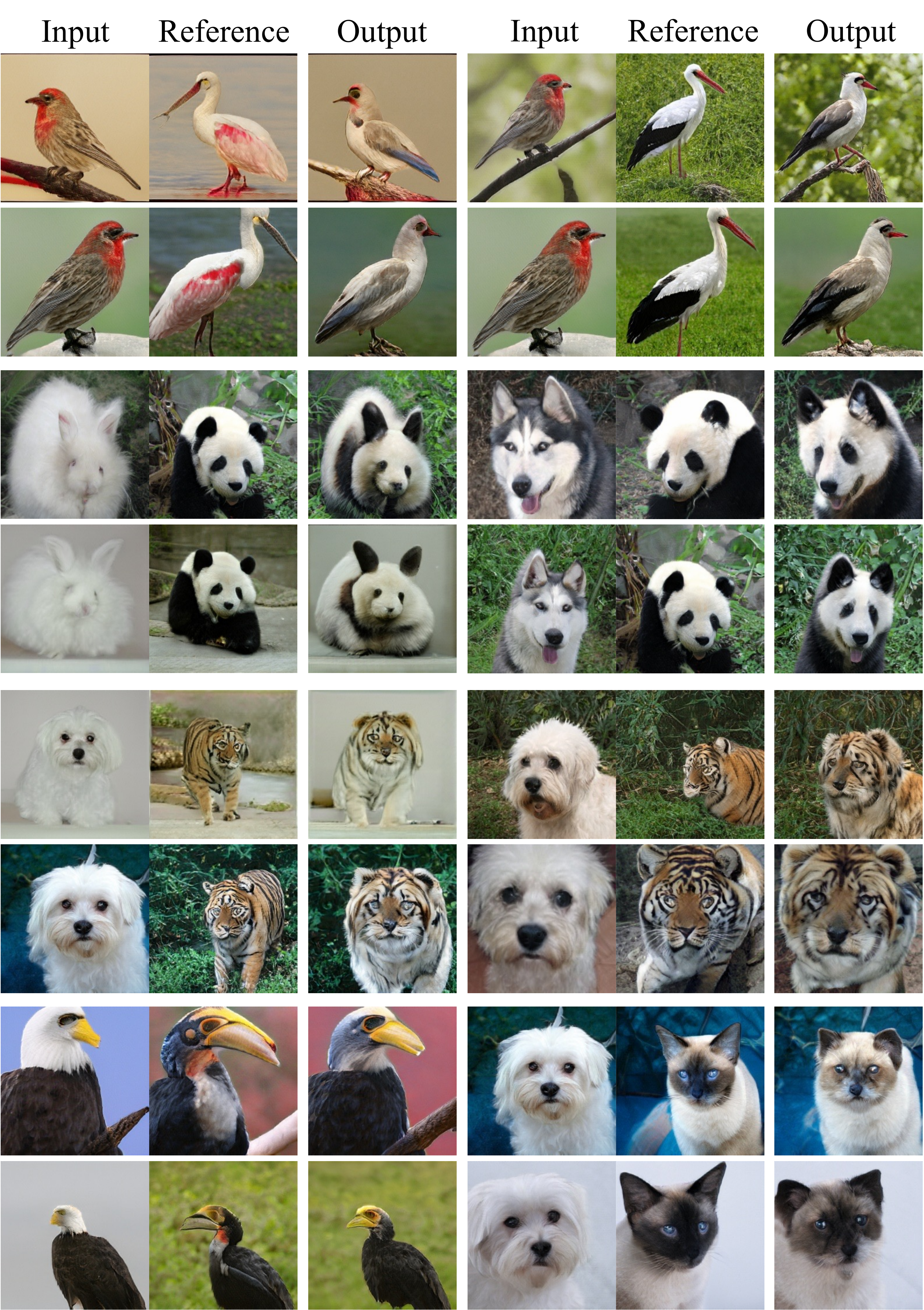}
    \vspace{-5pt}
    \caption{
        \textbf{Category hybridization} by mixing the channels that are relevant to two different categories.
        For each group, the first two columns present the original syntheses, while the third column shows the hybridization result, which successfully fuses the characteristics (including both shape and appearance) of both classes.
    }
    \label{appendix:fig:channel_mixing}
    \vspace{-10pt}
\end{figure*}

Given a target class, our \classaware can unsupervisedly identify channels responsible for the class-oriented attributes, based on which image editing can be achieved via altering a single channel. We provide results of more classes for category-oriented attribute editing in \cref{appendix:fig:attribute_editing}.
Given one input class and one reference class, mixing channels relevant for the two classes can harmoniously mix attributes from two classes and obtain the realistic output.
\cref{appendix:fig:channel_mixing} presents more results for category hybridization via channel mixing.

\begin{figure*}[t]
    \centering
    \includegraphics[width=0.85\linewidth]{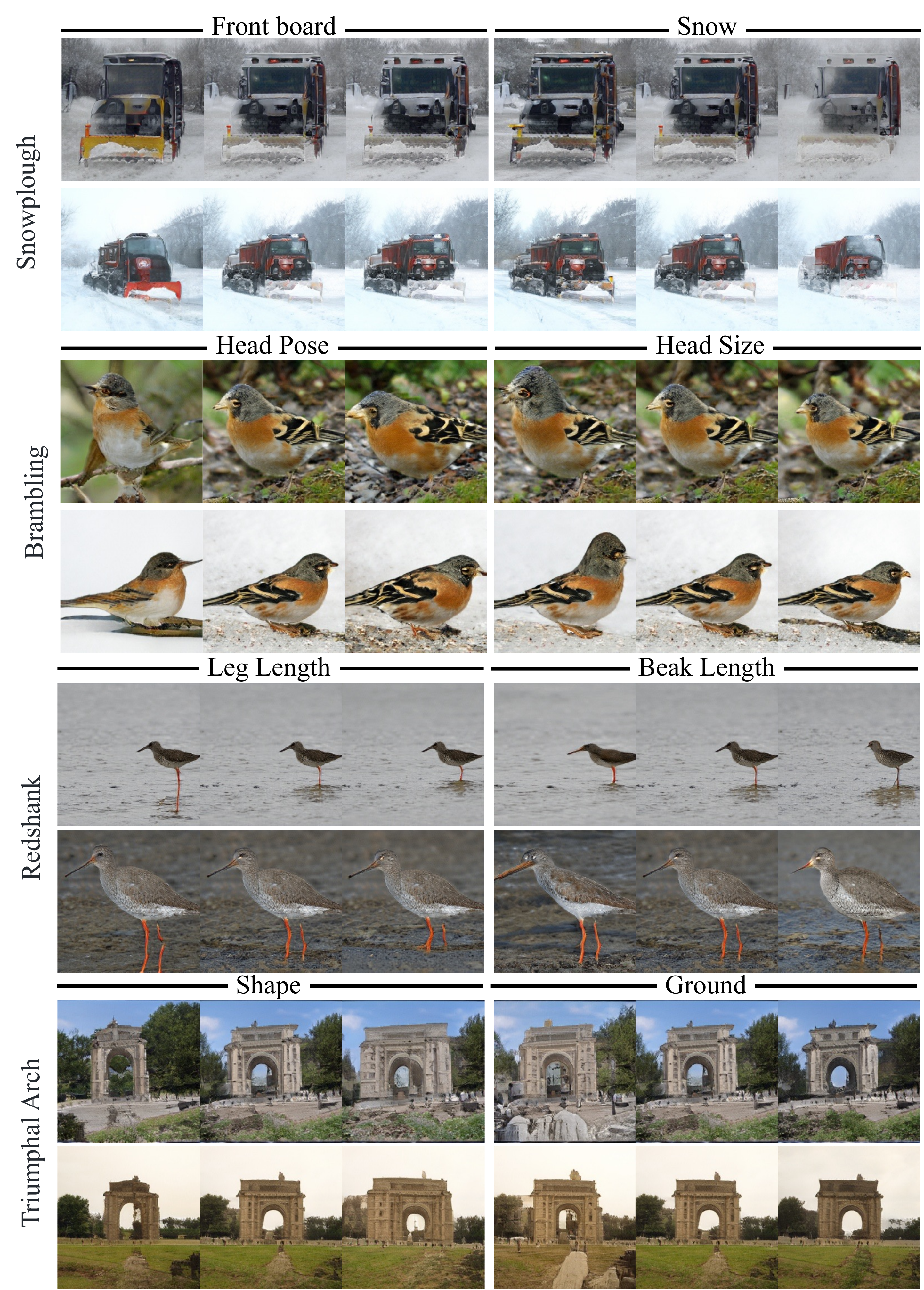}
    \vspace{-5pt}
    \caption{
        \textbf{Category-oriented attribute editing.}
        For each group of images, the middle one is input, while the left and the right one are manipulating the single channel along the positive and the negative direction.
        Corresponding class names are annotated on the left.
    }
    \label{appendix:fig:attribute_editing}
    \vspace{-10pt}
\end{figure*}

\end{document}